\documentclass{bmvc2k}

\usepackage{amsmath}
\usepackage{amssymb}
\usepackage{graphicx}
\usepackage{multirow}
\usepackage{booktabs}
\usepackage{caption}
\usepackage{subcaption}
\usepackage{algorithmic}
\usepackage{algorithm}
\usepackage[dvipsnames]{xcolor}

\title{Leveraging Multi-Modal Information to Enhance Dataset Distillation}

\addauthor{Zhe Li}{zhe.li@fau.de}{1}
\addauthor{Hadrien Reynaud}{hadrien.reynaud19@imperial.ac.uk}{2}
\addauthor{Bernhard Kainz}{b.kainz@imperial.ac.uk}{1,2}

\addinstitution{
 Department AIBE\\
 FAU Erlangen-Nürnberg\\
 Erlangen, Germany
}
\addinstitution{
 Department of Computing\\
 Imperial College London\\
 London, UK
}

\runninghead{Z Li, H Reynaud, B Kainz}{Multi-Modal Dataset Distillation}

\begin{document}

\maketitle

\begin{abstract}
Dataset distillation aims to create a small and highly representative synthetic dataset that preserves the essential information of a larger real dataset. Beyond reducing storage and computational costs, related approaches offer a promising avenue for privacy preservation in computer vision by eliminating the need to store or share sensitive real-world images. Existing methods focus solely on optimizing visual representations, overlooking the potential of multi-modal information.
In this work, we propose a multi-modal dataset distillation framework that incorporates two key enhancements: caption-guided supervision and object-centric masking. 
To leverage textual information, we introduce two strategies: caption concatenation, which fuses caption embeddings with visual features during classification, and caption matching, which enforces semantic alignment between real and synthetic data through a caption-based loss. To improve data utility and reduce unnecessary background noise, we employ segmentation masks to isolate target objects and introduce two novel losses: masked feature alignment and masked gradient matching, both aimed at promoting object-centric learning.
Extensive evaluations demonstrate that our approach improves downstream performance while promoting privacy protection by minimizing exposure to real data.
\end{abstract}

\section{Introduction}
\label{sec:intro}

Computer vision has rapidly advanced with the rise of deep learning, leading to remarkable achievements in tasks such as image classification, segmentation, and object detection. These breakthroughs have been largely fueled by the availability of large-scale datasets like ImageNet-1K. However, reliance on massive datasets comes at a significant computation and storage cost, which creates major challenges for efficient training and model deployment, especially in resource constrained environments.
To address these limitations, dataset distillation~\cite{wang2018dataset, such2020generative} has emerged as a promising solution. This technique aims to synthesize a small set of optimized samples that encapsulate the essential information of the full dataset. Models are expected to achieve competitive performance using far fewer synthetic data samples without privacy information.
 
Recent approaches of dataset distillation can be broadly categorized into two directions: matching-based methods~\cite{zhao2020dataset,zhao2021dataset,zhao2023DM,cazenavette2022dataset} and methods using generative models~\cite{cazenavette2023generalizing,su2024d}.
Matching-based methods optimize the alignment between real and synthetic images. In contrast, other approaches leverage pretrained GAN-based or diffusion models to assist the generation of high quality images.
While both techniques have demonstrated success in image classification, the potential benefits of multi-modal information have been overlooked. 
Recent works~\cite{xu2024low, wu2023vision} have explored image-text similarity objectives similar to vision language tasks. However, the direct utilization of multi-modal data on large-scale datasets remains unexplored. A major challenge is the lack of ground truth annotations for multi-modal information in most datasets.
To overcome this limitation, we utilize state-of-the-art methods~\cite{wang2024instancediffusion,zhang2024recognize,ren2024grounded,kirillov2023segment,li2023blip} to generate captions, segmentation masks and bounding boxes for real images. To the best of our knowledge, our approach is the first to comprehensively leverage multi-modal data, such as caption descriptions and object-centric masks, to the ImageNet-1K dataset for the purpose of dataset distillation.

Building on this insight, we propose a novel framework that integrates caption features and segmentation masks into the distillation pipeline. Specifically, we design two approaches for integrating caption features to provide high-level semantic information: caption feature concatenation, which directly concatenates caption embeddings with visual features, enabling the model to process linguistic and visual information jointly; and caption matching, which measures the alignment between captions of real and synthetic images alongside gradient matching.
Segmentation masks help localize important regions within images. To leverage this, we introduce masked gradient matching and masked distribution matching to enhance object-centric learning. By removing background regions, the model is guided to focus on salient object areas and mitigate the risk of overfitting to irrelevant background information.

Our contributions are as follows:
\begin{enumerate}
  \item We utilize multi-modal information in dataset distillation by incorporating caption features and segmentation masks to enhance feature representation. Due to the absence of ground-truth multi-modal annotations, we generate these annotations using pre-trained models.
  \item We propose two approaches for incorporating caption features: (i) Caption Concatenation: Caption features are concatenated with visual features before the classification stage, enriching semantic representations. (ii) Caption Matching: a caption matching loss is applied to align the captions of real and synthetic images.
  \item We design two methods that utilize segmentation masks, masked gradient matching and masked distribution matching, to learn object-specific features.
  \item Extensive experiments validate the effectiveness and generalization ability of our proposed methods, with improved performance observed across various data subsets.
\end{enumerate}

\section{Related work}

Following the initial model selection frameworks~\cite{wang2018dataset, such2020generative}, researchers have extensively explored matching-based methods designed to align the training dynamics on synthetic images with those trained on real datasets. Representative methods include Dataset Condensation with Gradient Matching (DC)~\cite{zhao2020dataset,zhao2021dataset,zhang2023accelerating}, Distribution Matching (DM)~\cite{zhao2023DM,zhao2023improved}, Matching Training Trajectories (MTT)~\cite{cazenavette2022dataset}, Sequential Subset Matching~\cite{du2023sequential}, and feature alignment approaches leveraging convolutional networks~\cite{wang2022cafe,sajedi2023datadam}.
In parallel, various alternative approaches have been proposed to enhance distillation quality, including factorization techniques~\cite{liu2022dataset}, methods that minimize accumulated trajectory errors~\cite{du2023minimizing}, calibration strategies~\cite{zhu2023rethinking}, and frequency domain optimizations~\cite{shin2023frequency}. To address challenges associated with high-frequency noise prevalent in pixel space, several methods~\cite{zhao2022synthesizing,cazenavette2023generalizing,su2024d,gu2024efficient} synthesize images in latent spaces, utilizing pretrained generative models.
While prior works have explored vision language methods~\cite{xu2024low, wu2023vision} on other datasets for dataset distillation. Our approach is the first to comprehensively leverage multi-modal data of the widely-used ImageNet-1K specifically for the dataset distillation process, demonstrating notable improvements over existing methods.


\section{Method}
\label{sec:formatting}

Dataset distillation refers to the process of compressing the rich information from a large real dataset into a significantly smaller synthetic dataset. The goal is to enable models trained on sythetic datasets to achieve comparable performance to those trained on the full dataset in downstream tasks such as classification.
Given a real dataset $\mathcal{T} = \{(x_i, y_i)\}_{i=1}^{N}$, where each image $x_i \in \mathbb{R}^{3 \times H \times W}$ is paired with a class label $y_i \in \{0, 1, 2, \ldots, C\}$, the dataset consists of $N$ samples across $C$ classes. The goal of dataset distillation is to synthesize $\text{IPC}$ (images per class) samples, resulting in a small synthetic dataset $\mathcal{S} = \{(s_i, y_i)\}_{i=1}^{N_s}$, where $N_s = \text{IPC} \times C$ and $N_s \ll N$. We generate multi-modal annotations for a large-scale dataset in Section~\ref{sec:multimodalgeneration}. Then, we present methods that incorporate caption features in Section~\ref{sec:captionincorporation} and segmentation masks in Section~\ref{sec:maskalignment}.

\subsection{Multi-modal annotation generation}
\label{sec:multimodalgeneration}

\begin{figure}[tb]
\begin{center}
    \includegraphics[width=0.8\linewidth]{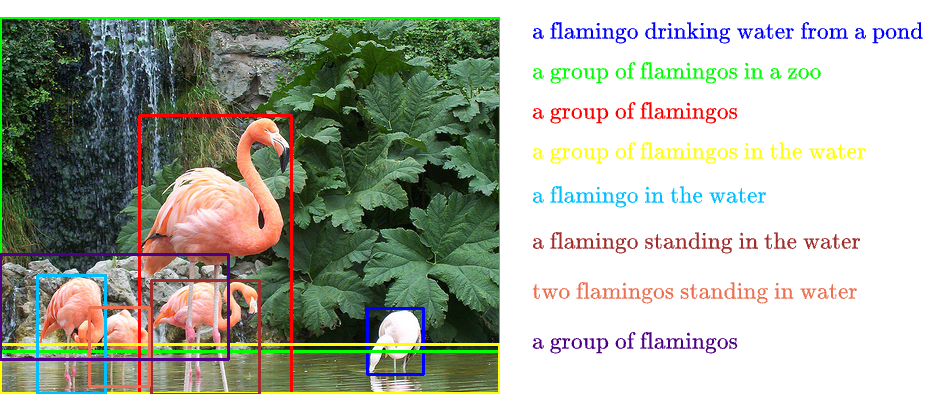}
\end{center}
\vspace{-3mm}
   \caption{Annotations of a sample from flamingo class.}
   \label{fig:annotations}
\vspace{-2mm}
\end{figure}

Integrating multi-modal data into vision tasks holds great promise, but it is often hindered by the absence of ground truth annotations in large-scale datasets. To overcome this limitation, we employ advanced pretrained models to automatically generate rich annotations, including captions, bounding boxes, and segmentation masks. These annotations provide the necessary semantic and spatial context to enable effective multi-modal learning. An example of the generated annotations is shown in Figure~\ref{fig:annotations}.

In detail, we use the Recognize Anything Model (RAM)~\cite{zhang2024recognize} to generate object labels for each image, thereby establishing a broad semantic understanding. As a powerful image-tagging model, RAM is capable of detecting and assigning multiple class labels to an image, enabling us to produce a rich set of labels that accurately describe the objects within each image.
Building on the output from RAM, we then generate bounding boxes for individual objects using Grounded Segment Anything (Grounded-SAM)~\cite{ren2024grounded}, a model that combines object grounding with segmentation capabilities. After identifying the bounding boxes, we further refine the segmentation by applying the Segment Anything Model (SAM)~\cite{kirillov2023segment}, which produces precise, pixel-level segmentation masks for each detected object.
To generate descriptive captions, we use BLIP-V2~\cite{li2023blip}, a state-of-the-art Vision-Language Model (VLM) trained on large-scale vision-language datasets. By feeding both the full image and the cropped object instances into BLIP-V2, we obtain an image-level caption as well as instance-level captions that describe each object in detail. These captions are subsequently encoded using CLIP~\cite{radford2021learning} to generate the corresponding caption features. We notice that preprocessing a large scale dataset as ImageNet-1K is time-consuming. However, the data is generated only once and stored, it can be reused directly in the distillation process.

\begin{figure*}[tb]
\begin{center}
   \includegraphics[width=1.0\linewidth, trim=0 7mm 0 5mm, clip]{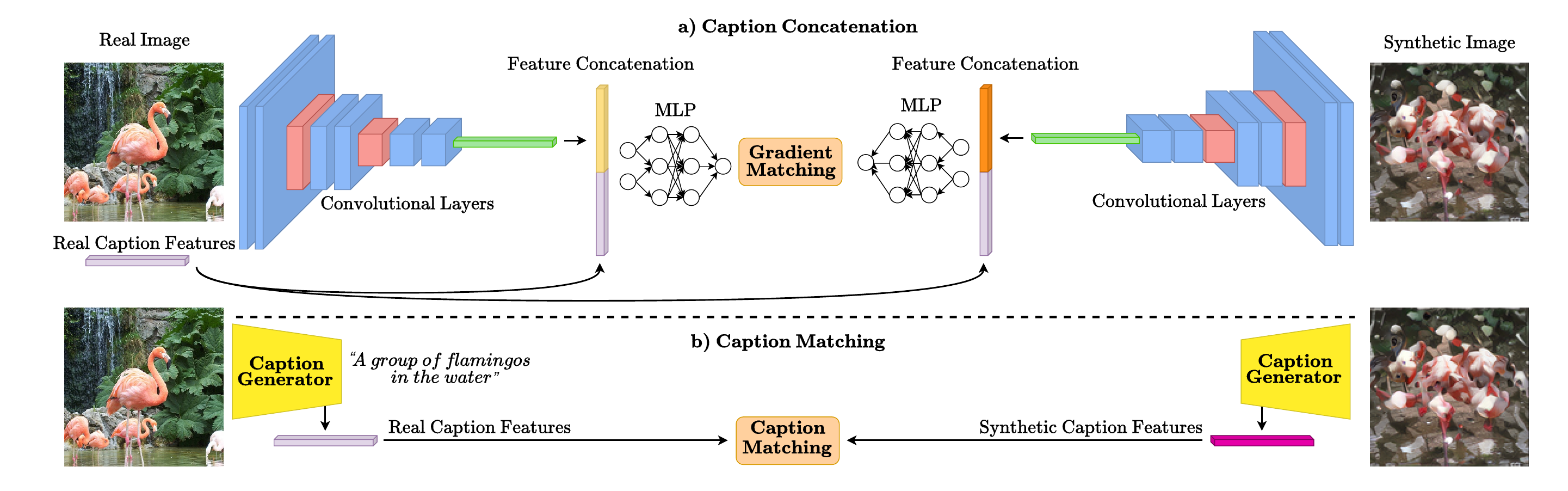}
\end{center}
\vspace{-3mm}
   \caption{Overview of the Caption Combination Framework. (a) Caption Concatenation: The caption feature is integrated with the image feature before being passed through the linear layer for probability prediction. (b) Caption Matching: In each iteration, caption features are extracted from synthetic images and aligned with those from real images.}
   \label{fig:captioncombination}
\vspace{-2mm}
\end{figure*}

\subsection{Caption incorporation}
\label{sec:captionincorporation}

Caption sentences convey semantic information about object classes, attributes, their relationships, and the contextual relevance between objects and their environment. Incorporating this information can enhance the quality of the distilled dataset by enabling the production of more meaningful and generalizable representations.
A key challenge is how to integrate caption features into the distillation process while preserving the stability and effectiveness of the optimization. We propose two distinct strategies: caption feature concatenation and caption matching. The overall framework is illustrated in Figure~\ref{fig:captioncombination}.

\paragraph{Caption feature concatenation.} In this approach, we directly concatenate caption embeddings and the feature representations obtained from the penultimate layer of the training model. This fused feature vector is then passed to the last layer of the model, which predicts class probabilities based on both visual and textual information, as in Figure~\ref{fig:captioncombination} a) \textit{Caption Concatenation}.
We adopt the framework of GLAD~\cite{cazenavette2023generalizing} as our backbone and extend it by introducing caption features before the classification stage while keeping the core distillation process unchanged. 
A key advantage of this approach is its computational efficiency. Since captions are generated in advance and introduced only at the classification stage, the distillation process itself remains largely unaffected. This makes the method easy to integrate into existing frameworks while still allowing models to benefit from additional textual context, potentially improving the discriminability of similar classes.

\paragraph{Caption matching.}
Current methods typically align gradients, distributions, or other representations between real and synthetic images. Building on this idea, we propose aligning caption features from real and synthetic images to further enhance semantic consistency.
To this end, we introduce the Caption Matching method, illustrated in Figure~\ref{fig:captioncombination} b) \textit{Caption Matching}.
For real images, caption features are precomputed and stored. During training, caption features for synthetic images are generated on-the-fly using the pretrained BLIP-V2~\cite{li2023blip} model.
Because the number of object-level captions may vary for synthetic images, we use only the full-image captions to ensure consistency across samples.
With caption features from both real and synthetic images, we compute a caption matching loss that enhance semantic alignment between them.
This loss is jointly optimized with the standard gradient matching loss. The final optimization objective is formulated as follows:
\begin{equation}
  \mathcal{L} = \mathcal{L}_{grad} + \lambda\mathcal{L}_{caption}
  \label{eq:captionmatchingloss}
\end{equation}
where \(\mathcal{L}_{grad}\) represents the traditional gradient matching loss, \(\mathcal{L}_{caption}\)  is the mean square error (MSE) that enforces similarity between real and synthetic caption features, and the hyperparameter \(\lambda\) are weighting factors.
This approach ensures that the distilled dataset aligns with the real dataset in both visual and linguistic characteristics. As a result, the synthetic samples become more semantically meaningful, leading to better generalization and improved performance in downstream tasks. This highlights the value of incorporating multi-modal consistency into the distillation process.

\begin{figure*}[tb]
\begin{center}
    \includegraphics[width=1.0\linewidth]{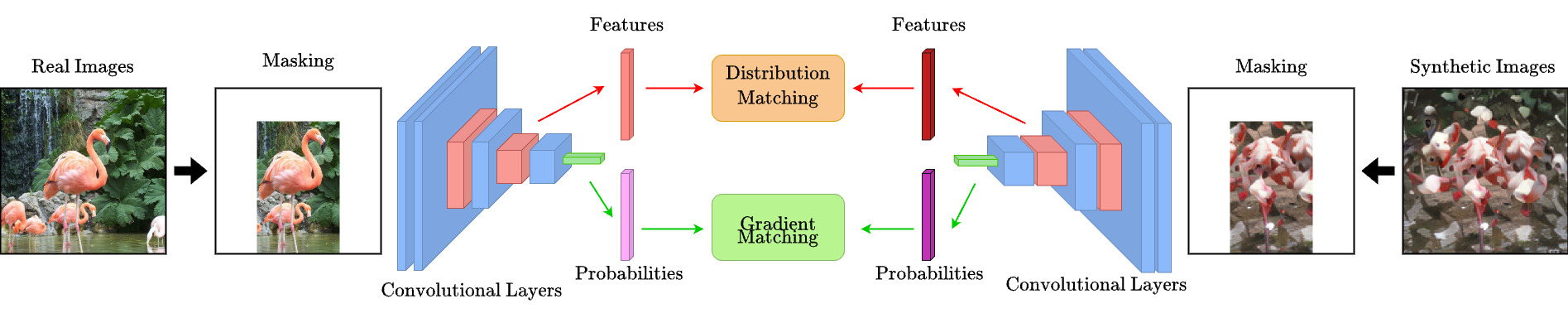}
\end{center}
\vspace{-3mm}
   \caption{Overview of the Mask Matching Framework.}
   \label{fig:maskmatching}
\vspace{-3mm}
\end{figure*}


\subsection{Object-centric alignment}
\label{sec:maskalignment}

Since dataset distillation aims to generate a highly representative dataset for each class, eliminating irrelevant or noisy background content is essential. To achieve this, we utilize generated segmentation masks to remove background regions from both real and synthetic images, ensuring the model focuses exclusively on the object of interest. This object-centric refinement enhances the model's ability to learn meaningful features and reduces the risk of overfitting to background noise. We introduce two strategies for incorporating masks: masked gradient matching and Masked distribution matching.

\paragraph{Masked gradient matching.}
Traditional gradient matching optimize synthetic images by aligning their gradient updates with those computed from real images. However, these methods do not explicitly account for object-centric differences, which can lead to suboptimal alignment, especially when real images contain varying background content. To mitigate this issue, we propose a masked gradient matching loss that focuses exclusively on foreground objects, thereby reducing the influence of background noise.
Since each segmentation mask corresponds to a single object, we begin by filtering the masks to retain only those that match the class currently being processed. Then, we calculate masked images. For real images, we apply each mask to its corresponding image. For synthetic images, we apply all relevant masks to each image, effectively expanding their dimensionality to match the masked real images for alignment.
Once masked images are obtained for both real and synthetic data, we perform gradient matching between them. 
Given a model parameterized \(\theta\), let \(\nabla_{\theta}\mathcal{L}(x, y)\) denote the gradient of the loss function with respect to an input image \(x\) and its label \(y\).
Our masked gradient matching objective is defined as:
\begin{equation}
  \mathcal{L}_{grad} = ||\nabla_{\theta}\mathcal{L}(\hat{x}_{real}, y) - \nabla_{\theta}\mathcal{L}(\hat{x}_{syn}, y)||^2
  \label{eq:important1}
\end{equation}
where both the \(\hat{x}_{real}\) and \(\hat{x}_{syn}\) are masked images. This formulation encourages alignment of object-focused gradients between real and synthetic data, promoting more semantically meaningful optimization during the distillation process.

\paragraph{Masked distribution matching.}
Traditional distribution matching aims to optimize synthetic images by aligning their intermediate-layer features with those extracted from real images. We extend this idea by applying distribution matching specifically to masked images, ensuring that the comparison focuses on object-centric features rather than background content.
let \(f_{\theta}\) denotes the model with parameters \(\theta\), and \(f_{\theta}(x)\) represent the feature vector extracted from an input image \(x\).
We compute the Mean Squared Error (MSE) loss between the features of masked real and synthetic images, as in Eq.~\ref{eq:maskeddmmseloss}:

\begin{equation}
  \mathcal{L}_{MSE} = \frac{1}{B}\sum_{i=1}^{B}\sum_{j=1}^{\text{IPC}}||f_{\theta}(\hat{x}_{real,i}) - f_{\theta}(\hat{x}_{syn,j})||^2
  \label{eq:maskeddmmseloss}
\end{equation}
where \(\hat{x}_{real,i}\) and \(\hat{x}_{syn,j}\) represent the background-masked real and synthetic images, respectively, and \(B\) is the batch size. This object-focused loss encourages the distilled dataset to retain meaningful, class-relevant features while minimizing the influence of irrelevant background information, leading to improved generalization in downstream tasks.

\section{Experiments}
\paragraph{Datasets and metrics.}
We evaluate our approach on subsets of the ImageNet-1K dataset~\cite{deng2009imagenet}. Specifically, we run experiments on \(10\) subsets of images at a resolution of \(128 \times 128\). Each subset contains \(10\) classes and each class consists of all \(1300\) images belonging to this class. Additionally, we also evaluate performance on \(5\) subsets at a higher resolution of \(256 \times 256\) to further assess the robustness and effectiveness of our method.
For a fair comparison with state-of-the-art methods, we report classification test accuracy across all experiments.

\paragraph{Implementation Details.}
For all experiments, we use GLAD~\cite{cazenavette2023generalizing} as the backbone model, while the distillation is performed using a ConvNet model. To comprehensively evaluate the generalization capability of the distilled dataset, we assess its performance across a diverse model pool comprising \(5\) classifiers: ConvNet, ResNet18, VGG11, ViT, and AlexNet. The hyperparameter \(\lambda\) in eqution~\ref{eq:captionmatchingloss} is set to \(0.2\) to balance the two losses to a similar scale.
We evaluate each of the \(5\) models for \(5\) times to ensure the robustness, and we report both the mean and standard deviation of the results. 

\begin{table}[tb]\setlength{\tabcolsep}{3pt}
   \begin{center}
   \resizebox{1.0\textwidth}{!}{%
      \begin{tabular}{lcccccccccc}
         \toprule
         & ImNet-A & ImNet-B & ImNet-C & ImNet-D & ImNet-E & ImNette & ImWoof & ImNet-Birds & ImNet-Fruits & ImNet-Cats \\
         \midrule
         DM~\cite{zhao2023DM} & 39.4\(_{\pm{1.8}}\) & 40.9\(_{\pm{1.7}}\) & 39.0\(_{\pm{1.3}}\) & 30.8\(_{\pm{0.9}}\) & 27.0\(_{\pm{0.8}}\) & 30.4\(_{\pm{2.7}}\) & 20.7\(_{\pm{1.0}}\) & 26.6\(_{\pm{2.6}}\) & 20.4\(_{\pm{1.9}}\) & 20.1\(_{\pm{1.2}}\) \\        
         GLAD(DM)~\cite{cazenavette2023generalizing} & 41.0\(_{\pm{1.5}}\) & 42.9\(_{\pm{1.9}}\) & 39.4\(_{\pm{0.7}}\) & 33.2\(_{\pm{1.4}}\) & 30.3\(_{\pm{1.3}}\) & 32.2\(_{\pm{1.7}}\) & 21.2\(_{\pm{1.5}}\) & 27.6\(_{\pm{1.9}}\) & 21.8\(_{\pm{1.8}}\) & 22.3\(_{\pm{1.6}}\) \\
         \midrule
         DC~\cite{zhao2020dataset} & 43.2\(_{\pm{0.6}}\) & 47.2\(_{\pm{0.7}}\) & 41.3\(_{\pm{0.7}}\) & 34.3\(_{\pm{1.5}}\) & 34.9\(_{\pm{1.5}}\) & 34.2\(_{\pm{1.7}}\) & 22.5\(_{\pm{1.0}}\) & 32.0\(_{\pm{1.5}}\) & 21.0\(_{\pm{0.9}}\) & 22.0\(_{\pm{0.6}}\)  \\
         GLAD (DC)~\cite{cazenavette2023generalizing} & 44.1$_{\pm{2.4}}$ & 49.2$_{\pm{1.1}}$ & 42.0$_{\pm{0.6}}$ & 35.6$_{\pm{0.9}}$ & 35.8$_{\pm{0.9}}$ & 35.4$_{\pm{1.2}}$ & 22.3$_{\pm{1.1}}$ & 33.8$_{\pm{0.9}}$ & 20.7$_{\pm{1.1}}$ & 22.6$_{\pm{0.8}}$\\ 
         \midrule
         Cap Cat (DC) & \textbf{46.5\(_{\pm{1.1}}\)} & 49.0\(_{\pm{0.8}}\) & \textbf{44.3\(_{\pm{1.0}}\)} & \textbf{36.9\(_{\pm{1.2}}\)} & \textbf{36.0\(_{\pm{0.9}}\)} & \textbf{36.5\(_{\pm{1.9}}\)} & 23.0\(_{\pm{0.9}}\) & \textbf{34.2\(_{\pm{1.6}}\)} & \textbf{22.6\(_{\pm{1.3}}\)} & \textbf{23.5\(_{\pm{1.4}}\)} \\
         
         Cap Match (DC) & 46.4\(_{\pm{0.8}}\) & 48.7\(_{\pm{0.4}}\) & 42.8\(_{\pm{1.0}}\) & 35.0\(_{\pm{1.7}}\) & 34.5\(_{\pm{1.1}}\) & 36.1\(_{\pm{1.2}}\) & 23.4\(_{\pm{0.7}}\) & 33.9\(_{\pm{1.2}}\) & 21.4\(_{\pm{1.5}}\) & 22.7\(_{\pm{1.0}}\) \\
         
         Masked DM (DC) & 45.9\(_{\pm{2.0}}\) & \textbf{50.0\(_{\pm{1.7}}\)} & 43.7\(_{\pm{1.7}}\) & 35.7\(_{\pm{1.4}}\) & 35.2\(_{\pm{0.9}}\) & 35.6\(_{\pm{0.7}}\) & 22.6\(_{\pm{1.1}}\) & 34.1\(_{\pm{1.3}}\) & 22.0\(_{\pm{1.1}}\) & 23.5\(_{\pm{0.9}}\) \\
     
         Masked DC (DC) & \textbf{46.5\(_{\pm{1.4}}\)} & 48.6\(_{\pm{1.6}}\) & 43.2\(_{\pm{0.2}}\) & 35.1\(_{\pm{2.0}}\) & 35.0\(_{\pm{1.2}}\) & 36.2\(_{\pm{1.6}}\) & \textbf{23.5\(_{\pm{1.0}}\)} & 33.4\(_{\pm{1.8}}\) & 21.8\(_{\pm{1.7}}\) & 22.3\(_{\pm{0.7}}\) \\
         
         \bottomrule
      \end{tabular}
    }
\end{center}
   \caption{Results for IPC$=1$ on subsets of ImageNet-1K at a resolution $128 \times 128$. Both distillation and classification are trained using the ConvNet. DC denotes Dataset Condensation, DM denotes distribution matching.}
   \label{tab:imagenetsubset128}
   \vspace{-3mm}
\end{table}

\subsection{Comparison with state-of-the-art}

Table~\ref{tab:imagenetsubset128} shows our results using both caption-based and mask-based methods compared to the state-of-the-art methods. We calculate the relative improvement as (B - A) / A.
The caption concatenation (\textit{Cap Cat (DC)}) approach improves accuracy by \(9.18\%\) (\(22.6\) vs. \(20.7\)) on ImNet-Fruits subset, demonstrating the benefits of leveraging semantic information from text descriptions.
The caption matching (\textit{Cap Match (DC)}) achieves up to \(5.22\%\) improvement across \(10\) subsets.
The masked distribution matching (\textit{Masked DM (DC)}) shows an increase in performance of \(6.28\%\) (\(22.0\) vs. \(20.7\)) on ImNet-Fruits subset.
The masked gradient matching (\textit{Masked DC (DC)}) yields a performance gain of up to \(5.44\%\) across all subsets.
Our experimental results indicate that leveraging additional multi-modal information enhances dataset distillation performance.

\begin{figure}[tb]
\begin{center}
     \begin{subfigure}[b]{0.48\columnwidth}
         \centering
         \includegraphics[width=\linewidth]{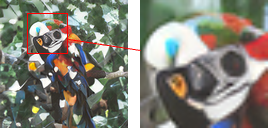}
         \caption{Caption concatenation}
         \label{}
     \end{subfigure}
     \hfill
     \begin{subfigure}[b]{0.48\columnwidth}
         \centering
         \includegraphics[width=\linewidth]{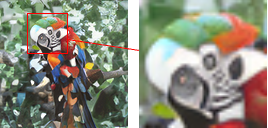}
         \caption{Caption matching}
         \label{}
     \end{subfigure}
     \hfill
     \begin{subfigure}[b]{0.32\columnwidth}
         \centering
         \includegraphics[width=\linewidth]{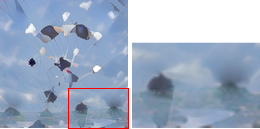}
         \caption{Caption concatenation}
         \label{}
     \end{subfigure}
     \hfill
     \begin{subfigure}[b]{0.32\columnwidth}
         \centering
         \includegraphics[width=\linewidth]{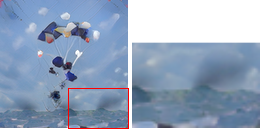}
         \caption{Masked DM}
         \label{}
     \end{subfigure}
     \hfill
     \begin{subfigure}[b]{0.32\columnwidth}
         \centering
         \includegraphics[width=\linewidth]{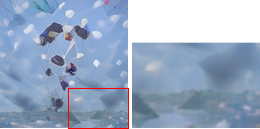}
         \caption{Masked DC}
         \label{}
     \end{subfigure}

\end{center}
\vspace{-3mm}
   \caption{Qualitative results of different methods. (a) Macaw from ImNet-Birds, generated using caption concatenation. (b) The same Macaw class, generated using caption matching. (c), (d), and (e) show Parachute from ImNette, where mask-based methods effectively reduce background elements.}
\label{fig:qualitative128}
\end{figure}

Figure~\ref{fig:qualitative128} shows the qualitative results at \(128\times 128\) resolution.
Figure~\ref{fig:qualitative128} (a) and (b) show results generated with caption information integrated. We zoom in on the head of the Macaw to highlight class-specific characteristics.
Figure~\ref{fig:qualitative128} (c) serves as a baseline for comparison with mask-based methods, using caption concatenation.
Figure~\ref{fig:qualitative128} (d) and (e) show results generated using masked images. Object boundaries are enhanced while background artifacts are effectively suppressed, especially in the bottom regions highlighted by the red box. 
This demonstrates that the masking mechanism successfully reduces irrelevant gradients and guides the model to focus on object-specific features.

\begin{table*}[tb]\setlength{\tabcolsep}{3pt}
\begin{center}
   \resizebox{1.0\textwidth}{!}{%
      \begin{tabular}{lcccccccccc}
         \toprule
         & ImNet-A & ImNet-B & ImNet-C & ImNet-D & ImNet-E & ImNette & ImWoof & ImNet-Birds & ImNet-Fruits & ImNet-Cats \\
         \midrule
         DM~\cite{zhao2023DM} & 27.2$_{\pm{1.2}}$ & 24.4$_{\pm{1.1}}$ & 23.0$_{\pm{1.4}}$ & 18.4$_{\pm{1.7}}$ &17.7$_{\pm{0.9}}$ &20.6$_{\pm{0.7}}$ & 14.5$_{\pm{0.9}}$ & 17.8$_{\pm{0.8}}$ & 14.5$_{\pm{1.1}}$ & 14.0$_{\pm{1.1}}$\\
         GLAD(DM)~\cite{cazenavette2023generalizing} & 31.6$_{\pm{1.4}}$ & 31.3$_{\pm{3.9}}$ & 26.9$_{\pm{1.2}}$ & 21.5$_{\pm{1.0}}$ & 20.4$_{\pm{0.8}}$ & 21.9$_{\pm{1.1}}$ & 15.2$_{\pm{0.9}}$ & 18.2$_{\pm{1.0}}$ &20.4$_{\pm{1.6}}$ & 16.1$_{\pm{0.7}}$ \\
         \midrule
         DC~\cite{zhao2020dataset} & 38.7$_{\pm{4.2}}$ & 38.7$_{\pm{1.0}}$ & 33.3$_{\pm{1.9}}$ & 26.4$_{\pm{1.1}}$ &  27.4$_{\pm{0.9}}$ & 28.2$_{\pm{1.4}}$ & 17.4$_{\pm{1.2}}$ & 28.5$_{\pm{1.4}}$ & 20.4$_{\pm{1.5}}$ & 19.8$_{\pm{0.9}}$\\      
         GLAD(DC)~\cite{cazenavette2023generalizing} & 41.8$_{\pm{1.7}}$ & 42.1$_{\pm{1.2}}$ & 35.8$_{\pm{1.4}}$ & 28.0$_{\pm{0.8}}$ & 29.3$_{\pm{1.3}}$ & 31.0$_{\pm{1.6}}$ & 17.8$_{\pm{1.1}}$ & 29.1$_{\pm{1.0}}$ & 22.3$_{\pm{1.6}}$ & 21.2$_{\pm{1.4}}$ \\
         \midrule         
         Cap Cat (DC) & \textbf{43.4\(_{\pm{1.3}}\)} & 43.0\(_{\pm{1.4}}\) & 37.0\(_{\pm{1.1}}\) & 29.4\(_{\pm{1.3}}\) & 30.3\(_{\pm{1.4}}\) & 32.8\(_{\pm{1.8}}\) & 19.3\(_{\pm{1.0}}\) & 30.1\(_{\pm{1.0}}\) & \textbf{23.5\(_{\pm{1.1}}\)} & 20.8\(_{\pm{1.1}}\) \\
         Cap Match (DC) & 42.9\(_{\pm{1.1}}\) & 43.3\(_{\pm{1.1}}\) & \textbf{37.8\(_{\pm{1.1}}\)} & 29.0\(_{\pm{1.3}}\) & 30.7\(_{\pm{1.5}}\) & 32.9\(_{\pm{1.0}}\) & 19.4\(_{\pm{0.6}}\) & 29.7\(_{\pm{1.1}}\) & 23.1\(_{\pm{1.3}}\) & 21.4\(_{\pm{1.0}}\) \\
         Masked DM (DC) & 42.7\(_{\pm{1.8}}\) & \textbf{43.5\(_{\pm{1.5}}\)} & 37.3\(_{\pm{1.3}}\) & \textbf{30.1\(_{\pm{1.0}}\)} & 31.0\(_{\pm{1.3}}\) & \textbf{33.0\(_{\pm{1.3}}\)} & 19.3\(_{\pm{1.4}}\) & \textbf{30.5\(_{\pm{1.2}}\)} & 23.3\(_{\pm{1.0}}\) & 21.2\(_{\pm{0.8}}\)  \\
         Masked DC (DC) & 42.4\(_{\pm{1.4}}\) & 42.6\(_{\pm{1.1}}\) & \textbf{37.8\(_{\pm{0.2}}\)} & 29.4\(_{\pm{1.2}}\) & \textbf{31.7\(_{\pm{1.2}}\)} & 32.8\(_{\pm{1.0}}\) & \textbf{19.5\(_{\pm{1.4}}\)} & 30.3\(_{\pm{1.0}}\) & 22.8\(_{\pm{1.7}}\) & \textbf{21.6\(_{\pm{0.8}}\)}  \\
         \bottomrule
      \end{tabular}
    }
 \end{center}
   \caption{Cross Architecture Results for IPC\(=1\) at a resolution $128 \times 128$.}
   \label{tab:imagenetsubsetcross128}
   \vspace{-3mm}
\end{table*}

\subsection{Cross architecture results}

We conduct cross-architecture evaluations to assess the generalization ability of our approach in Table~\ref{tab:imagenetsubsetcross128}. The distillation phase is trained on a ConvNet, while the classification phase is evaluated on four different architectures: ResNet18, VGG11, ViT, and AlexNet.
Caption feature concatenation (Cap Cat) achieves \(8\%\) improvement (\(19.3\) vs. \(17.8\)) on the ImWoof subset, showing that enriching visual representations with semantic information improves classification performance.
Caption matching (Cap Match) yields \(9\%\) improvement (\(19.4\) vs. \(17.8\)) on ImWoof subset, highlighting the effectiveness of aligning caption semantics with real images during distillation.
Masked distribution matching (Masked DM) enhances performance by \(8\%\) (\(30.1\) vs. \(28.0\)) on the ImNet-D subset, demonstrating the effectiveness of focusing on object-centric features.
Similarly, masked gradient matching (Masked DC) achieves \(10\%\) improvement (\(19.5\) vs. \(17.8\)) on the ImWoof subset, indicating that restricting gradient updates to salient object regions enhances feature learning.

\subsection{Ablation study}

\begin{table}[tb]\setlength{\tabcolsep}{5pt}
\begin{center}
   \resizebox{0.7\columnwidth}{!}{%
      \begin{tabular}{lccccc}
         \toprule
          & ImNet-A & ImNet-B & ImNet-C & ImNet-D & ImNet-E  \\
         \midrule
         DC~\cite{zhao2020dataset} & 38.3$_{\pm{4.7}}$ & 32.8$_{\pm{4.1}}$ & 27.6$_{\pm{3.3}}$ & 25.5$_{\pm{1.2}}$ & 23.5$_{\pm{2.4}}$ \\
         GLaD~\cite{cazenavette2023generalizing} & 37.4$_{\pm{5.5}}$ & 41.5$_{\pm{1.2}}$ & 35.7$_{\pm{4.0}}$ & 27.9$_{\pm{1.0}}$ & 29.3$_{\pm{1.2}}$ \\
        
         \midrule         
         Cap Match (DC) & \textbf{44.1\(_{\pm{1.2}}\)} & \textbf{43.1\(_{\pm{1.4}}\)}  & 36.9\(_{\pm{1.4}}\) & 29.4\(_{\pm{1.5}}\) & 30.0\(_{\pm{1.5}}\) \\
         Masked DC (DC) & 43.5\(_{\pm{1.5}}\)  & 42.0\(_{\pm{1.1}}\) &  \textbf{37.4\(_{\pm{1.3}}\)} & \textbf{29.7\(_{\pm{0.9}}\)} & \textbf{31.3\(_{\pm{1.1}}\)} \\
         \midrule    
         Cap ConvNet &  44.0\(_{\pm{1.0}}\) & 47.4\(_{\pm{0.7}}\)  & 41.1\(_{\pm{0.7}}\) & 33.2\(_{\pm{0.9}}\) & 33.7\(_{\pm{1.1}}\) \\   
         Masked ConvNet &  45.2\(_{\pm{2.1}}\) & 48.1\(_{\pm{1.1}}\)  & 41.5\(_{\pm{0.9}}\) & 33.0\(_{\pm{1.7}}\) & 33.0\(_{\pm{1.5}}\) \\
         \bottomrule
      \end{tabular}
    }
\end{center}
   \caption{Cross-architecture results for IPC\(=1\) at a resolution $256 \times 256$.}
   \label{tab:imagenetsubset256}
\end{table}

\begin{figure*}[t]
\begin{center}
     \begin{subfigure}[b]{0.48\columnwidth}
         \centering
         \includegraphics[width=\linewidth]{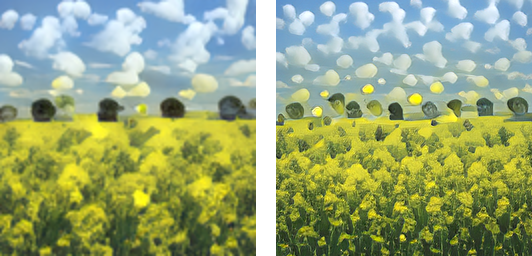}
         \caption{Caption matching}
         \label{}
     \end{subfigure}
     \hfill
     \begin{subfigure}[b]{0.48\columnwidth}
         \centering
         \includegraphics[width=\linewidth]{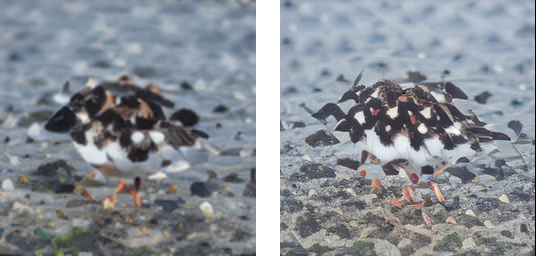}
         \caption{Masked gradient matching}
         \label{}
     \end{subfigure}

\end{center}
\vspace{-3mm}
   \caption{Qualitative results at different resolutions. (a) Rapeseed from ImNet-A, with the left image at \(128\times 128\) resolution and the right image at \(256\times 256\). (b) Ruddy Turnstone from ImNet-B at the same setting.}
\label{fig:qualitative256}
\vspace{-3mm}
\end{figure*}

We extend our evaluation to higher-resolution images (\(256\times256\)), with results in Table~\ref{tab:imagenetsubset256}. Increasing image resolution introduces additional challenges, such as greater computational complexity and more fine-grained details to capture.
Our approach consistently improves performance across all \(5\) subsets in cross-architecture evaluations.
The caption matching method (Cap Match) demonstrates a notable improvement of \(17.9\%\) (\(44.1\) vs. \(37.4\)) on the ImNet-A subset. 
The masked gradient matching method (Masked DC) achieves an improvement of \(6.8\%\) (\(31.3\) vs. \(29.3\)) on the ImNet-E subset.
This improvement confirms that the proposed approach has better generalization to complex images.
The results from Cap ConvNet and Masked ConvNet show higher performance when both distillation and classification use ConvNet model.
Figure~\ref{fig:qualitative256} shows qualitative comparisons between two image resolutions: \(128\times 128\) and \(256\times 256\). 
The higher-resolution images exhibit finer object details, such as the increased number of clouds in the sky and the clearer body parts of the Ruddy Turnstone.

\section{Conclusion}
In this work, we integrate caption-based supervision and leverage object-centric masking matching in dataset distillation. Captions provide rich semantic context that can complement visual features, and we propose two distinct approaches for incorporating them into the distillation process. The caption concatenation enables models to leverage linguistic information without altering the distillation process. The caption matching ensures that synthetic images maintain semantic consistency with real images.
Moreover, we propose two object-centric methods for dataset distillation by isolating target objects and remove background distractions. Masked distribution matching enforces consistency at the feature representation level. Masked gradient matching optimizes learning dynamics of the training model.
By eliminating irrelevant background details, our approach ensures that models focus on object features.
Comprehensive evaluations demonstrate that integrating caption-based guidance or object-centric masks leads to superior performance on downstream classification tasks.

\paragraph{Acknowledgments:} HPC resources were provided by the Erlangen National High Performance Computing Center (NHR@FAU), under the NHR projects b143dc and b180dc. NHR is funded by federal and Bavarian state authorities, and NHR@FAU hardware is partially funded by the DFG - 440719683. H.R. was supported by Ultromics Ltd., the UKRI Centre for Doctoral Training in Artificial Intelligence for Healthcare  (EP / S023283/1). We acknowledge the use of Isambard-AI National AI Research Resource (AIRR)~\cite{mcintosh2024isambard}. Isambard-AI is operated by the University of Bristol and is funded by the UK Government’s DSIT via UKRI; and the Science and Technology Facilities Council [ST/AIRR/I-A-I/1023]. The authors received funding from the ERC-project MIA-NORMAL 101083647,  DFG 513220538, 512819079, and by the state of Bavaria (HTA).

\bibliography{egbib}
\end{document}